\documentclass[conference]{IEEEtran}
\pdfoutput=1
\usepackage{blindtext}

\usepackage{times}
\usepackage{epsfig}
\usepackage{graphicx}
\usepackage{amsmath}
\usepackage{amssymb}

\usepackage{caption}
\usepackage{subcaption}

\usepackage{algpseudocode}
\usepackage{algorithm}

\usepackage{tabu}

\usepackage{multirow}
\usepackage{colortbl}
\definecolor{kugray5}{RGB}{224,224,224}

\bstctlcite{IEEEexample:BSTcontrol}

\title{Multi-modal Tracking for Object based SLAM}

\author{
        Prateek Singhal, 
        Ruffin White, 
        Henrik Christensen\\
        Institute of Robotics and Intelligent Machines\\ 
        Georgia Institute of Technology, \\
        Atlanta, Georgia, USA
}

\begin{document}

\maketitle

\begin{abstract}
We present an on-line 3D visual object tracking framework for monocular cameras by incorporating spatial knowledge and uncertainty from semantic mapping along with high frequency measurements from visual odometry. Using a combination of vision and odometry that are tightly integrated we can increase the overall performance of object based tracking for semantic mapping. We present a framework for integration of the two data-sources into a coherent framework through information based fusion/arbitration. We demonstrate the framework in the context of OmniMapper\cite{trevor14:omnimapper} and present results on 6 challenging sequences over multiple objects compared to data obtained from a motion capture systems. We are able to achieve a mean error of 0.23m for per frame tracking showing 9$\%$ relative error less than state of the art tracker.    
\end{abstract}


\section{Introduction}
\label{sec:introduction}

Current SLAM systems operate at either the level of sparse features using edges or salient points \cite{klein2007:PTAM}\cite{PTAM2} in the scene or do dense tracking of the whole scene \cite{newcombe2011kinectfusion}. We propose a method which integrates the semantic information of model based tracking and mapping with a sparse feature based tracking. Most recent methods try to use features which can be tracked over long periods of time. We go one level further to include objects in the map as semantic entities which can be effectively tracked over long periods. Our approach allows us to handle complex motions of the camera in a cluttered scene while preventing the map from growing too large or discrepancies developing in camera trajectory.  

\begin{figure}
        \begin{subfigure}[b]{0.5\linewidth}
                \includegraphics
                    [width=\linewidth,height = 5cm]
                    {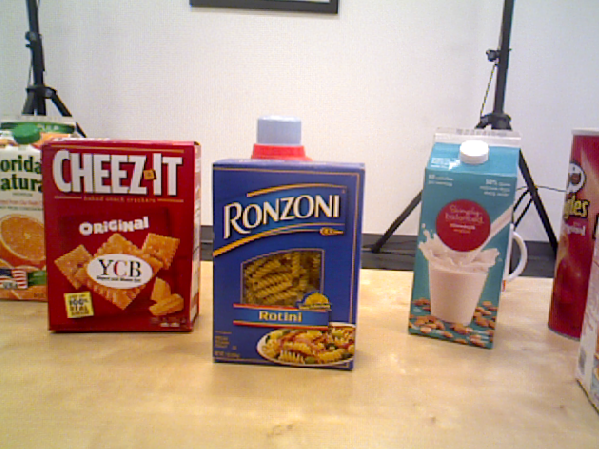}
        \end{subfigure}%
        \begin{subfigure}[b]{0.5\linewidth}
                \includegraphics
                    [width=\linewidth,height = 5cm ]
                    {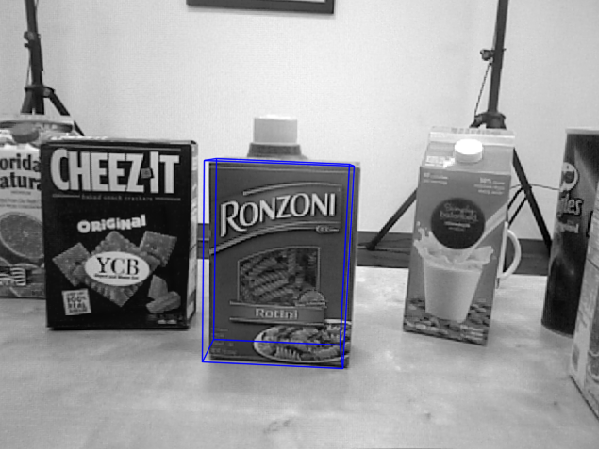}
        \end{subfigure}%
        \quad
        \begin{subfigure}[b]{0.5\linewidth}
                \includegraphics
                    [width=\linewidth,height = 5cm ]
                    {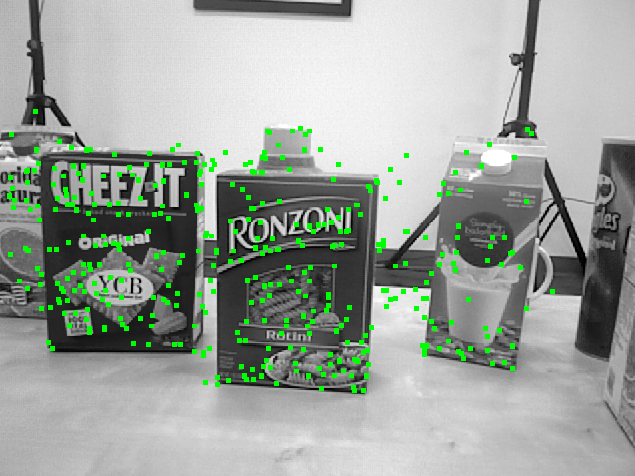}
        \end{subfigure}%
        \begin{subfigure}[b]{0.5\linewidth}
                \includegraphics
                    [width=\linewidth,height = 5cm ]
                    {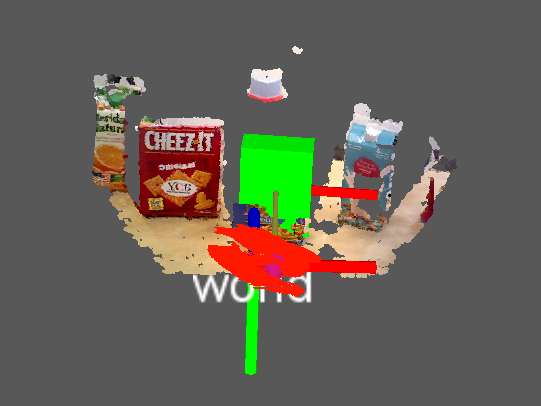}
        \end{subfigure}
    \caption{3D model based visual tracking aided by SLAM. \textbf{(Clockwise from top left)} Original image input, the model rendered on image using pose from model based tracker, the optimized estimates for the pose and landmarks in the world frame and sparse features used for visual odometry. (Point cloud serves for visualization only.)}
    \label{fig:factorgraph}
\end{figure}

Recently, there has been a growing interest in incorporating semantics into building maps. SLAM methods typically operate without the knowledge of semantics and generally fall into the trap of being unscalable with time, while purely detection based methods from images rely solely on the image features without any knowledge of the 3D scene, leading to longer processing times. We exploit the relationship of objects in the scene to build a map which is scalable over time as well as allowing detection to benefit from the knowledge of the current scene to provide an accurate pose. Some of the recent works like SLAM++ \cite{MorenoNSKD13} and SSFM \cite{bao2011semantic}  have approached this problem in a similar manner. SLAM++ uses a RGBD sensor to model the scene and the camera pose with known objects. SSFM tries to solve the 2 view problem using the relationship between points, objects and the camera in the scene. 

\begin{figure*}
    \setlength{\abovecaptionskip}{10pt plus 3pt minus 2pt} 
    \centering
    \includegraphics[page=2,
        width=\textwidth,
        trim=0 0 0 205,
        clip]
        {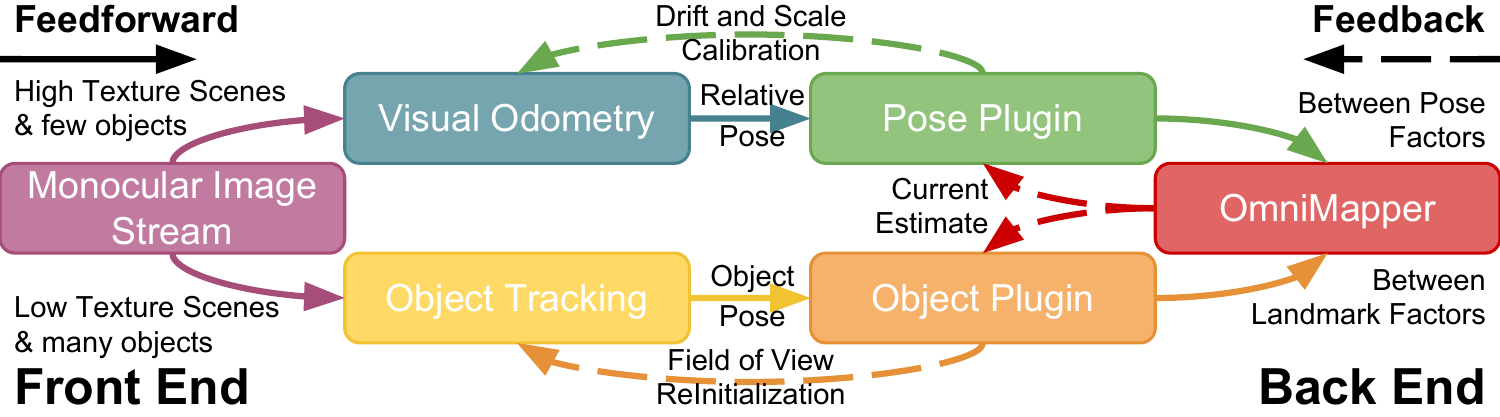}
    \caption{Flow diagram for proposed tracking framework. Each plugin process measurements from the same image, but commit different levels of information into the graph. With this form of cascading feedback, both plugins influence each other as they supervise their own measurement source's performance}
    \label{flowdiagram}
\end{figure*}

Various approaches detailing the use of geometric constraints over objects have been proposed in the past. Kim et al. build a 3D map and apply geometric constraints on the objects in the map. We use objects to construct the map and the trajectory of the camera, not as an ad hoc constraint for refinement of the map. Object tracking of the 6DOF pose of the object \cite{krull2014:6dof} \cite{lu2014:unsupervised}  has attracted lot of attention with the advent of depth sensors. These methods generally try to segment the image using level sets into contours, with 3D-3D and 2D-2D contour matching to estimate the pose. We use a state of the art object tracker which allows us to estimate the pose from a monocular image and show how using the current camera estimates from odometry improves both tracking and detection.    

We show our approach to be online on a CPU with incremental addition of pose and objects to the graph. Incremental operation on a CPU with monocular cameras has applications in real world robotics, like grasping in an industry setting or household setting, allowing for continuous object tracking even when the object goes out of the frame and comes back in, a common scenario with robot arms. Most of the methods in this field try to use AR markers to do efficient localization while our approach allows for work in marker free environments.

Our approach is comprised of a 3D model based object tracking and a visual odometry coupled in a factor graph to enable pose graph optimization. We assume a knowledge of the object's presence in the scene and characterize it by a 6DOF pose, while using salient edges of the object in the image to track it. A visual odometry algorithm tracking sparse features provides us with the current camera pose. Both of these measurements are complimentary in nature allowing us to deal with failure cases of both object tracking and SLAM. The object tracking allows us to scale the odometry to the metric scale and allows us to correct for drift in the odometry while the odometry provides reinitialization for object detection, when the object is lost, using past estimates. We demonstrate visual tracking and mapping in the presence of partial or full occlusion of the objects. We compare the results to data captured from a motion capture system. We have not been able to find a dataset for comparative evaluation. 

The outline of the paper is as follows : In Section \ref{sec:related}, we discuss some of the relevant methods in the literature. An overview of the system is given in Section \ref{sec:overview}. We describe our approach in Section \ref{sec:approach} and discuss our experimental results in Section \ref{sec:results}. Our approach is open source and is implemented in ROS, details of which are presented in Section \ref{sec:implementation}. We conclude the paper in Section \ref{sec:conclusions} by stating the goals we achieved and the possible future work in this direction.

\section{Related Work}
\label{sec:related}

Semantic Structure from Motion is a relatively new field without an exhaustive literature. Here we elaborate on two methods, SLAM++ Moreno et al. \cite{MorenoNSKD13}, and SSFM from Bao and Savarese \cite{bao2011semantic}, which have approached the same problem from different angles. Both of these methods, as we do, use objects as a semantic representation. They also exploit the structure of the factor graph to do efficient inference.

SLAM++: proposes an object oriented SLAM with a factor graph formulation using a RGBD sensor. They estimate the pose between the current camera pose and the object using a model based ICP, with the assumption that objects present in the scene are known. We make a similar assumption in our work. The odometry between the camera poses is found using  ICP. One of the major emphasis of the paper is on model based ICP for estimating the pose of the object in the current frame. This method suffers from the problem of large pre-processing of the point cloud in the current camera frame to find the correct model estimate from the database. In spirit our work is similar to theirs but our method uses a monocular camera with sparse tracking. 

SSFM: Semantic Structure from Motion approaches the traditional structure from motion problem by incorporating the semantic priors of objects in the formulation. A state of the art object detector is used to detect the bounding boxes of the object in the image and then standard CAD models are used to predict the scale and the pose probability maps. The points in the structure from motion are used to provide the camera pose. This method, though more generic than ours and SLAM++, requires more computationally extensive processing times due to class specific detection rather than instance specific detection. Furthermore, variations in the class can lead to failure of this system to estimate the pose of the object in the frame. Our take on object recognition for tracking needs to be instance specific rather than class specific favoring finer pose measurements over a large number of poses.  

Our method is comprised of Object Tracking and SLAM and explores approaches from both these fields. Object tracking has been one of the successes of computer vision, yet the larger body of the literature dealing with 2d tracking in images using bounding boxes is often limiting for robotics or Virtual Reality applications. Krull et al. \cite{krull2014:6dof} have proposed a 6DOF pose object RGBD tracker using a random forest based classifier for the pose of objects using depth maps. Lu and Gabe \cite{lu2014:unsupervised} recently proposed a dense object discovery, detection, tracking and reconstruction method using a RGBD sensor. Their method segments the image using level sets into contours and then uses 3D-3D and 2D-2D contour matching to estimate the pose of the object.      

Real-time visual SLAM systems calculate the visual odometry by tracking the 3d points over images. Semi-direct Visual Odomtery (SVO) \cite{forster14:svo} is a recent approach which uses an image alignment based method to track the features across images. It uses a depth based Gaussian filtering which allows it to converge to the correct 3d position. Here we use SVO, but make some modifications to initialize scaling up to metric scale. Large-Scale Direct(LSD) SLAM \cite{LSDslam} is another recent approach which utilizes the change in the gradient to estimate the odometry and the map. 

\section{System Overview}
\label{sec:overview}

  We show the outline of our system in Fig \ref{flowdiagram}. Monocular images are given as inputs to both tracking and visual odometry in real time. As each algorithm operates at different frame rates, object tracking performance varying with the number of tracked objects in the current scene, the measurements are published asynchronously. Omnimapper allows us to create a factor graph online with such asynchronous measurements. The data association per pose is evaluated based on an uncertainty gating criterion, wrong data associations are found and rectifying feedback is given to the tracker. Based on the severity of the data association failure, the current solution may be used to quickly reinitialise the algorithm, or if the current estimate is highly uncertain or unstable, an event signal may be used to trigger the algorithm's own reset procedure. In this way, information is shared across the algorithms in order to arrive at a superior consensus.

\section{Approach}
\label{sec:approach}

\begin{figure}
        \centering
        ~ 
            \includegraphics[page=3,
                scale=0.25,
                trim=100 75 100 75,
                clip]
                {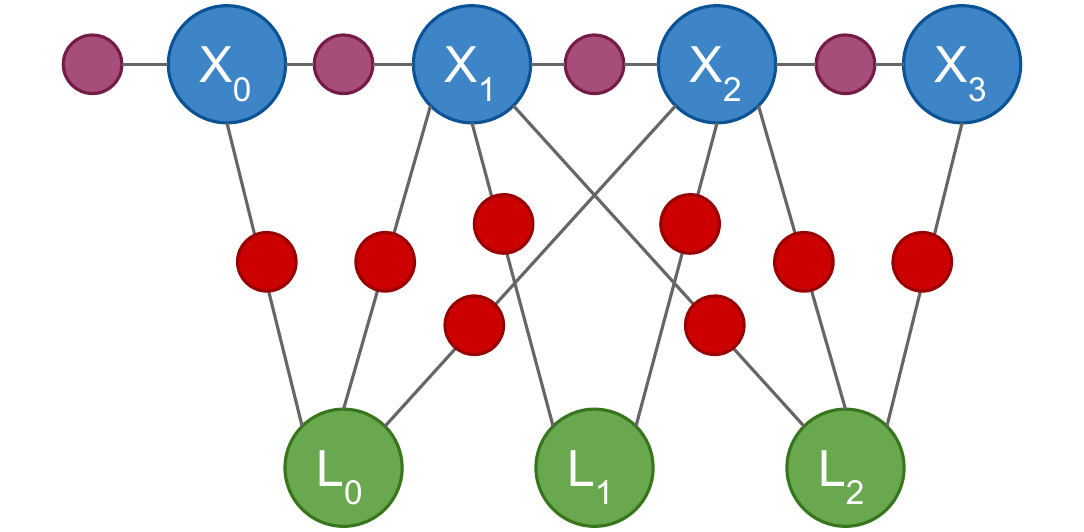}
            \caption{The pose of the camera at $i^{th}$ time step is $x_{i}$ with $i\in0\dots M$,
            a landmark is $l_{j}$ with $j\in1\dots N$ and a measurement is $z_{k}$,
            with $k\in1\dots K$. Factor graph representation. Blue circles denote camera poses ($X$) and green circles denote landmarks ($L$). Small purple circles represent odometry constraints and red circles represent landmark-pose constraint($Z$).}
        ~ 
        \label{fig:factorgraph}
\end{figure}

We formulate our problem as a factor graph as shown in Fig 3. Here, the objects are used as landmarks while a pose represents the current camera location along the trajectory. The object tracker provides the factor between the landmark and the pose, while the factor between consecutive poses is provided by visual odometry. This framework is generic and can incorporate multiple modalities of measurements represented as factors; other geometric features such as points and planes as landmarks can also be expressed as factors \cite{trevor14:omnimapper}. In this work, we limit the modality of landmarks to that of pose measurements of objects found in the current scene.

Our approach in detail is described as follows: 


\subsection{Object Recognition and Pose Estimation}

We use SURF features to model the objects in our database, as SURF is rotationally invariant compared to other faster features detectors allowing us to reinitialize robustly while tracking. During training each object in the database is stored with SURF feature descriptors, points and its corresponding 3D points of the model. During testing, we run a SURF descriptor on the image to recognize the objects in the scene. We build a kd-tree for each object in the database which is then matched using FLANN with the SURF descriptors in the current image. Matched keypoints of each object are matched with its corresponding 3D points in the database to obtain the pose of the object in the current frame using Perspective-n-Point \cite{epnp} algorithm. This process is carried out in a RANSAC framework to provide a robust estimate of the pose.     

\subsection{Visual Tracking}

Our method of visual tracking builds upon the 3D model-based visual tracking approach demonstrated in the work of Choi and Christensen \cite{choi12:robus_euclid}. This method of edge based tracking provides the requisite degree of flexibility and speed required for maintaining a high frequency estimate of the camera's pose relative to the object reference frame. By utilizing the pose obtained from object recognition to initialize the pose of the object model, we can project the 3D model onto the image using the camera's intrinsic and distortion parameters known a priori. The projected 3D model is rendered using only the salient edges within the original polygon mesh, as identifying sharp edges in the object will help to match edge features in the image. Once a hypothesis is initialized, the same edge-based measurement likelihood as illustrated in \cite{choi12:robus_euclid} is employed to guide a particle filter on the SE(3) group in tracking the object in the scene. 

The particle filter begins in much the same way, where our density function  $p(\textbf{X}_t|\textbf{Z}_{1:x})$is a set of weighted particles
\begin{equation}
\textbf{\textit{S}}_t = \{ (\textbf{X}^{(1)}_t,\pi^{(1)}_t),\dots,(\textbf{X}^{(N)}_t,\pi^{(N)}_t)\}
\end{equation}
With $\textbf{X}^{(n)}_t$ representing the sample of the current state $\textbf{X}_t$ in SE3, and $\pi^{(n)}_t$ the normalized weight proportional to the underlining likelihood function $p(\textbf{Z}_{1:x}|\textbf{X}_t)$, with $N$ the total number of particles. The likelihood of each particle is determined by the residual error between the projected model and the extracted edges from the image. The residual error is generated from sampling points along the model edges at a given density; thereafter, a 1D search along the direction orthogonal to the model edge to the nearest edge pixel is measured.
Normally a large number of particles is necessary for robust tracking, but to reduce the computation, a fewer number of particles is used and these are locally optimized using iteratively reweighted least squares (IRLS) \cite{choi10:realtime}.

We use the trackers' measure of its health based on the ratio between the valid visible sampled points in the image and sampled points on the image to determine whether the tracker should be reset or not. A sampled point along the edge is deemed valid if the error along the perpendicular direction is less than the maximum error. Here we set 32 pixels as the maximum error possible.

Multiple instances of such trackers can then be instantiated with respect to what objects are expected to be visible in the current environment. This also employs the successful trackers to aid poorer performing instances in static environments, and measured transformations from one object can help narrow the hypotheses of a second object temporarily lost from view. This feedback is described when expanding upon the flow of information within the proposed framework.

\subsection{Monocular Visual Odometry}

We use SVO for monocular visual odometry as it is open source and works in real-time. For making this paper self-sufficient, we explain it briefly. It follows the trend of having 2 parallel threads of mapping and tracking. 

SVO uses a Sparse Model-based Image Alignment between consecutive poses while refining the photometric error. This is aligned globally with features in the closest keyframes.
 \begin{equation}
\delta I(T,u) = I_k(\pi(T.\pi^{-1}(u,d_{u}))) - I_{k-1}(u) \qquad \forall  u\in R
\end{equation}

where T is the pose of the camera, $\pi$ is the mapping from the 3d point $d_u$ to the image feature u. R represents the region for which the depth of the features is known. This equation can be minimized iteratively to find the current camera pose with respect to the previous camera ($T_{k,k-1}$). The inverse depth of the features is estimated using recursive Gaussian filtering on the points along the epipolar lines using known pose. 

We initialize SVO using our object database based pose estimation at first keyframe; this process is repeated on the second keyframe to obtain a scaled estimate of the pose between the 2 keyframes. In the presence of multiple objects we chose the object with the highest number of inliers to intialize the keyframes.  We take care while initializing that the first 2 keyframes have a minimum disparity between them. The initialization formulation can be written as :

\begin{equation}
T_{2,1}^{c} = inv(T_{1}^{c,o})*T_{2}^{c,o}
\end{equation}

where $T_{2,1}$ is the pose between the first 2 keyframes for intialization,$T_{i}^{c,o}$ are the pose between the object and keyframes obtained from the object detection.

\begin{algorithm}
  \caption{Object Measurement Data Association ($O$)}

  \begin{algorithmic}
    \State $I_{k}$ = Image at $k^{th}$ frame
    \State counter = 0 
    \State init = true
    \While{$I_{k}$}
        \State $T_{k}^{c,o}$  = EBT($I_{k}$,init)  
        \State $T_{k,k-1}$ = SVO($I_{k}$)
        \If{(H($T_{k}^{c,o}$) = good)}
            \If{ Res $\leq$ $\alpha*(\sum)_{(T)^{w,o}}$ }
                \State Add $(T_{k}^{c,o})_{f}$ 
                \State counter = 0
                \State init = false
            \Else 
                \State Feedback()
            \EndIf
        \Else
            \State Feedback()
        \EndIf    
    \EndWhile
    \Function{feedback}{}
    \If{counter $\leq$ th}
        \State Reset EBT($(T_{k}^{c,o})_{e}$)
        \State counter++
        \State init = false
    \Else
        \State init = true
        \State Reset EBT(SURF)
    \EndIf
    \EndFunction
  \end{algorithmic}
\end{algorithm}

\subsection{Pose Graph Optimization}

Both visual tracking and visual odometry are combined together to form the complete factor graph. The factor graph constraints at time step k can be written down as :

\begin{equation}
T_{k}^{c,o} = T_{k}^{c,w}*T^{w,o}
\end{equation}
\begin{equation}
T_{k,k-1}^{c} = T_{k-1}^{c,w}*(T_{k}^{w,c})
\end{equation}

where $T_{k}^{c,o}$ is the pose of the object in the current camera frame, $T_{k}^{c,w}$ is the pose of the camera in the current frame to the world frame, $T^{w,o}$ is the pose of the object in the world frame, $T_{k,k-1}^{c}$ is the pose of the camera between time steps k and k-1 in $(k-1)^{th}$ camera frame. 

The energy to be minimized then can be written as:

\begin{equation}
E = min ||\sum_{k,o}((T_{k}^{c,o}-(T_{k}^{c,o})_{m}) +(T_{k,k-1}-(T_{k,k-1})_{m}))||_{2}   
\end{equation}

$(T_{k}^{c,o})_{m}$ is obtained from EBT, $(T_{k,k-1}^{c})_{m}$ is obtained from SVO, both of them are treated as measurements in the factor graph. The factor graph is optimized for $T_{k}^{c,w}$ and $T^{w,o}$ over all the objects and the poses.      

We use GTSAM \cite{gtsam} to optimize the factor graph at each time step. Since this is an online process, we use ISAM2 \cite{Kaess12ijrr} to estimate the current pose of the camera and the object. 

\subsection{Information gain based Feedback }
\label{sec:Feedback}

Visual tracking often suffers from failure due to fast motion or occlusion leading to addition of erroneous measurements in the factor graph. This situation is analogous to adding wrong loop closure constraints in traditional SLAM systems. We use Sequential Compatibility Nearest Neighbour (SCNN) \cite{jcbb} algorithm to reject erroneous data association. This can be formulated as :

\begin{equation*}   
    Res = ( (T)^{w,o}-(T_{k}^{w,c}*(T_{k}^{c,o})_{m})\\
\end{equation*}   
\begin{equation*}    
\hspace{12mm} 
(T_{k}^{c,o})_{f} =   
\begin{cases}
  (T_{k}^{c,o})_{m}, \quad\text{if  Res} \leq \alpha*(\sum)_{(T)^{w,o}} \\
  \text{NULL},\quad  \text{otherwise} 
\end{cases}
\end{equation*}

where Res is the residual between the current estimate of object in the world frame $(T)^{w,o}$ and the estimate from the current object tracking measurement $(T_{k}^{c,o})_{m}$ transformed to the world frame using the current estimate of the camera in the world frame $T_{k}^{w,c}$, $(T_{k}^{c,o})_{f}$ is the factor to be added and $\sum_{(T)^{w,o}}$ is the marginal covariance of the object in the world frame. 

If the measurement is rejected we reset the object tracker with the current estimated pose of the object which can be written as:
\begin{equation}
(T_{k}^{c,o})_{e} = T_{k}^{c,w}*T^{w,o};
\end{equation}
 
$(T_{k}^{c,o})_{e}$ is the predicted pose of the object in the $k_{th}$ camera frame. The method keeps resetting the pose of the object and increments a counter. When this counter overflows it triggers reinitialization of object tracking using SURF features from the object database. This is done as odometry accumulates drift and seldom leads to erroneous measurements of the predicted pose. The threshold \textit{th} for the counter is set to 500 in our case.

 We show the flow of our algorithm in Algorithm 1. EBT(SURF) corresponds to resetting object tracking using SURF, H($T_{k}^{c,o}$) is the function showing the status of the tracker. The $\alpha$ is the confidence of the object we would like our measurements to lie within, which is set to 2 in our case.

\section{Implementation}
\label{sec:implementation}

\begin{figure*}
        \begin{subfigure}[b]{0.33\linewidth}
                \includegraphics
                    [width=\linewidth]
                    {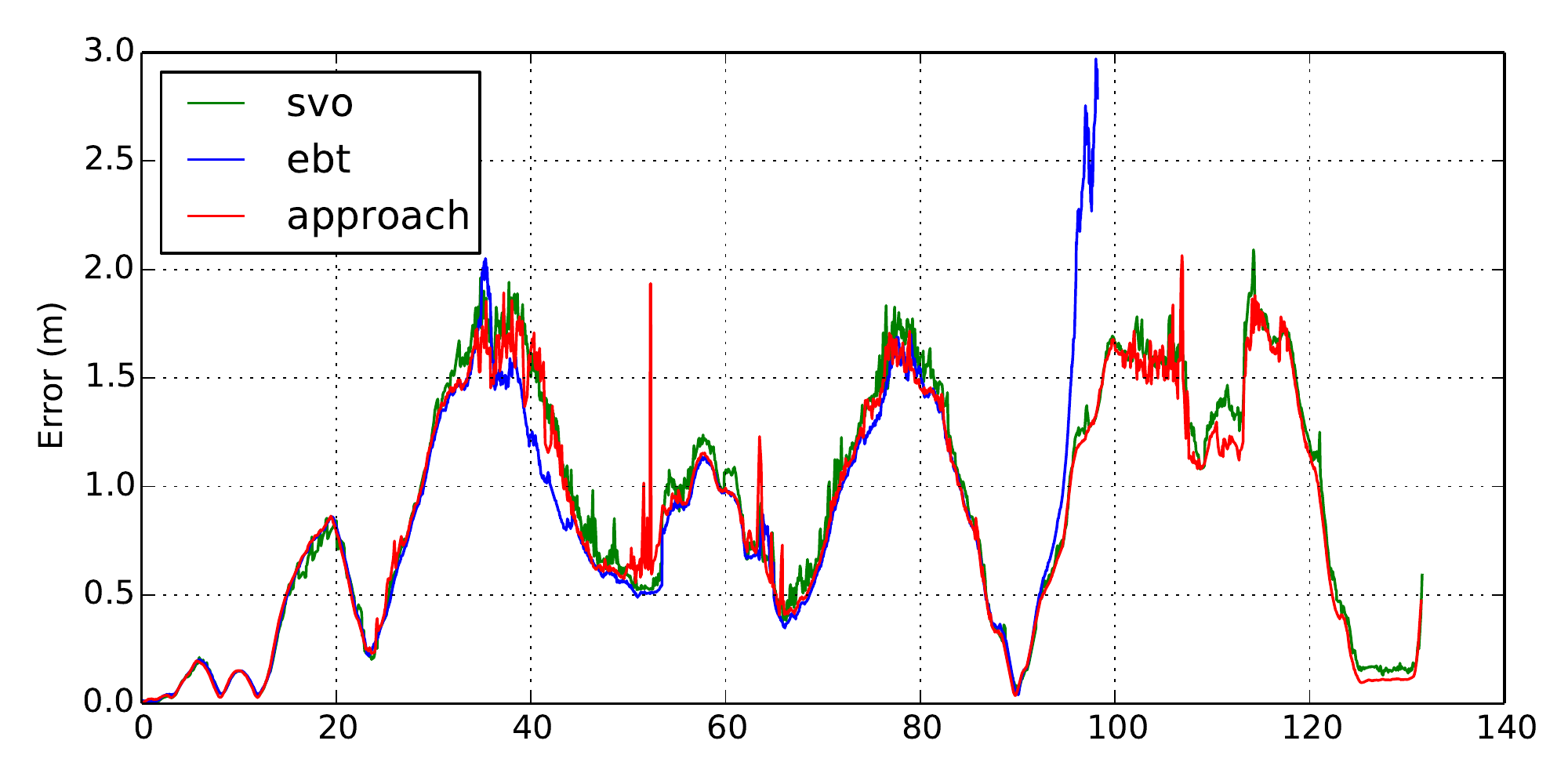}
        \end{subfigure}
        \begin{subfigure}[b]{0.33\linewidth}
                \includegraphics
                    [width=\linewidth]
                    {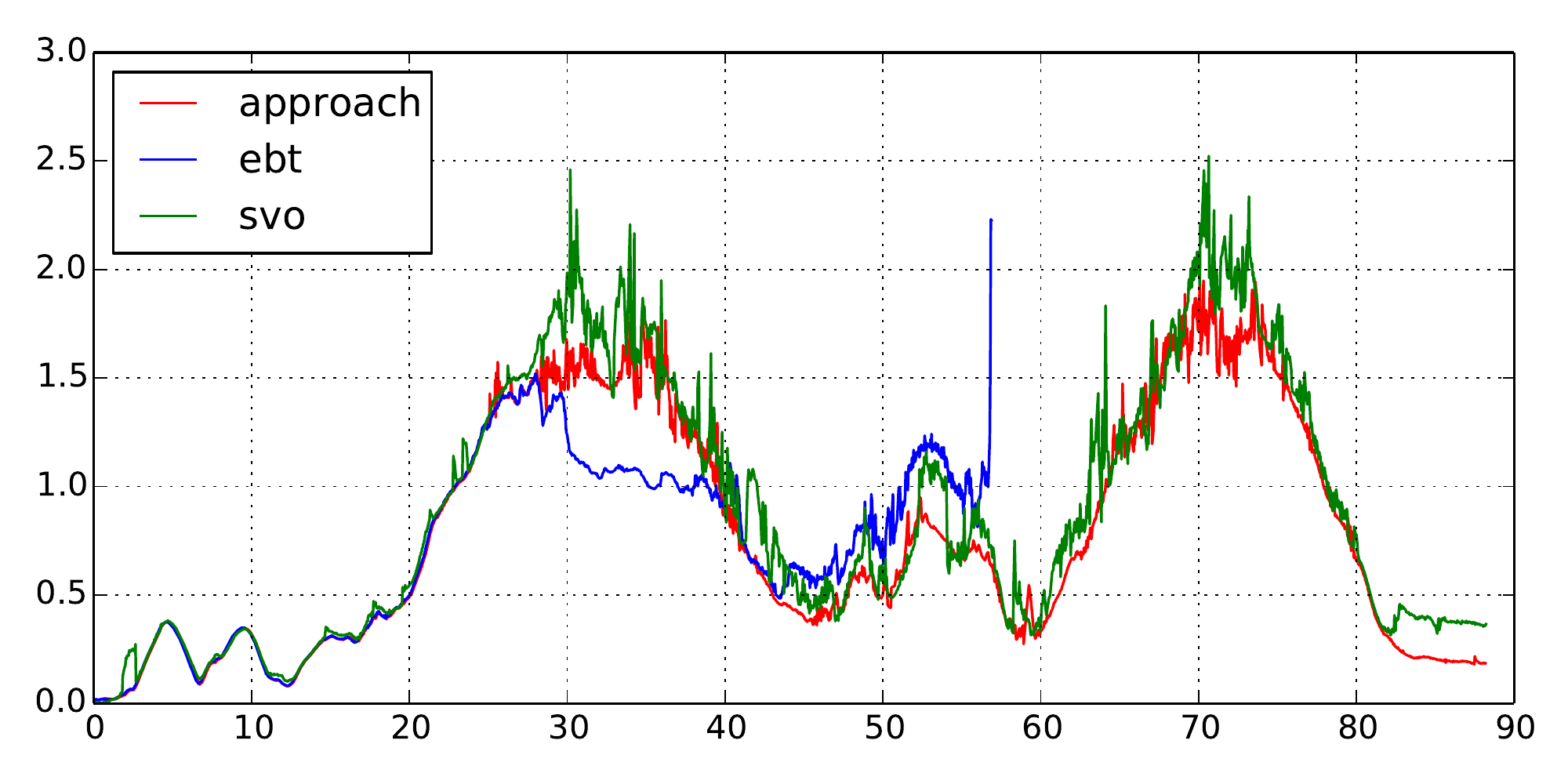}
        \end{subfigure}
        \begin{subfigure}[b]{0.33\linewidth}
                \includegraphics
                    [width=\linewidth]
                    {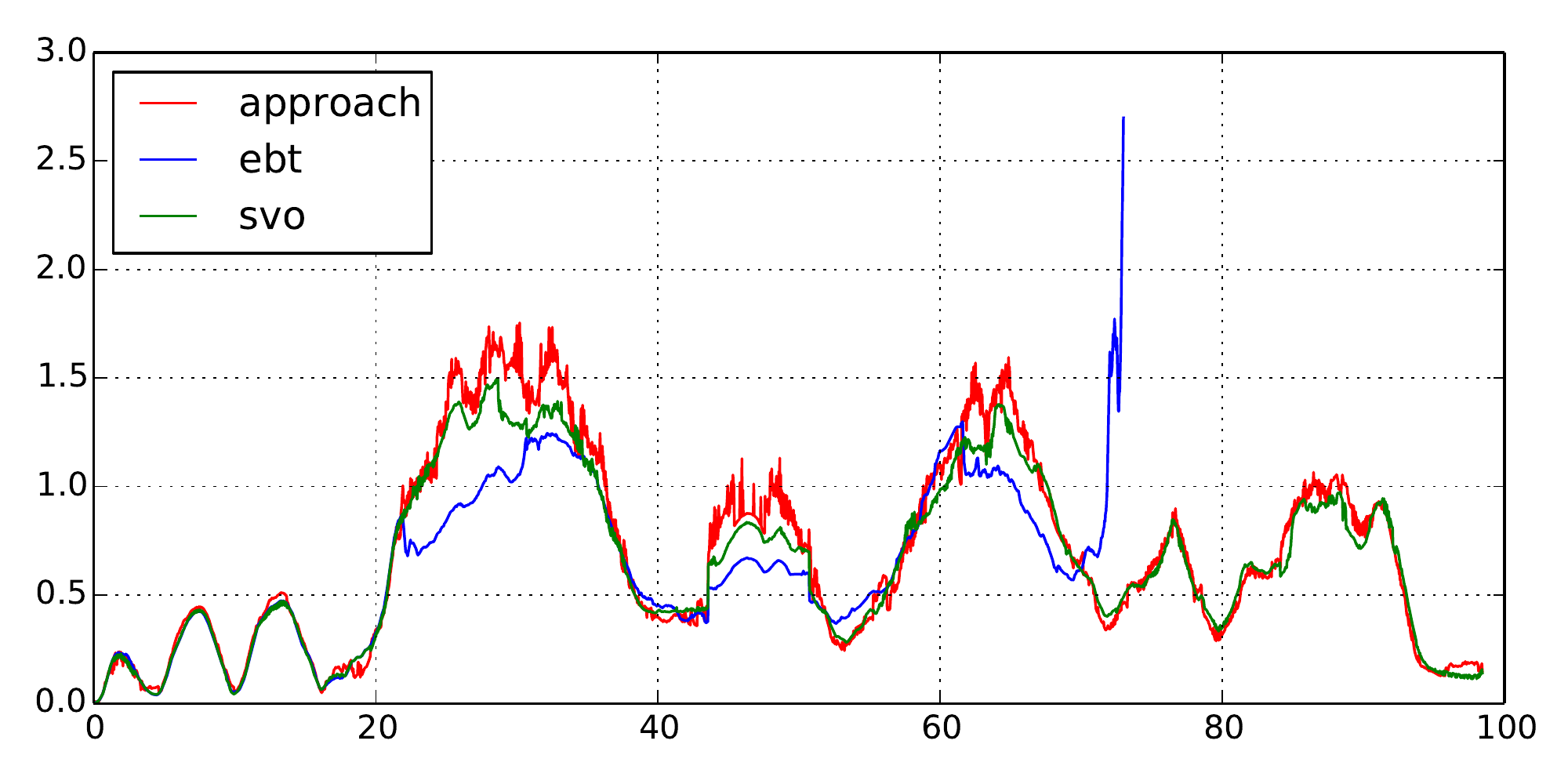}
        \end{subfigure}
        \begin{subfigure}[b]{0.33\linewidth}
                \includegraphics
                    [width=\linewidth]
                    {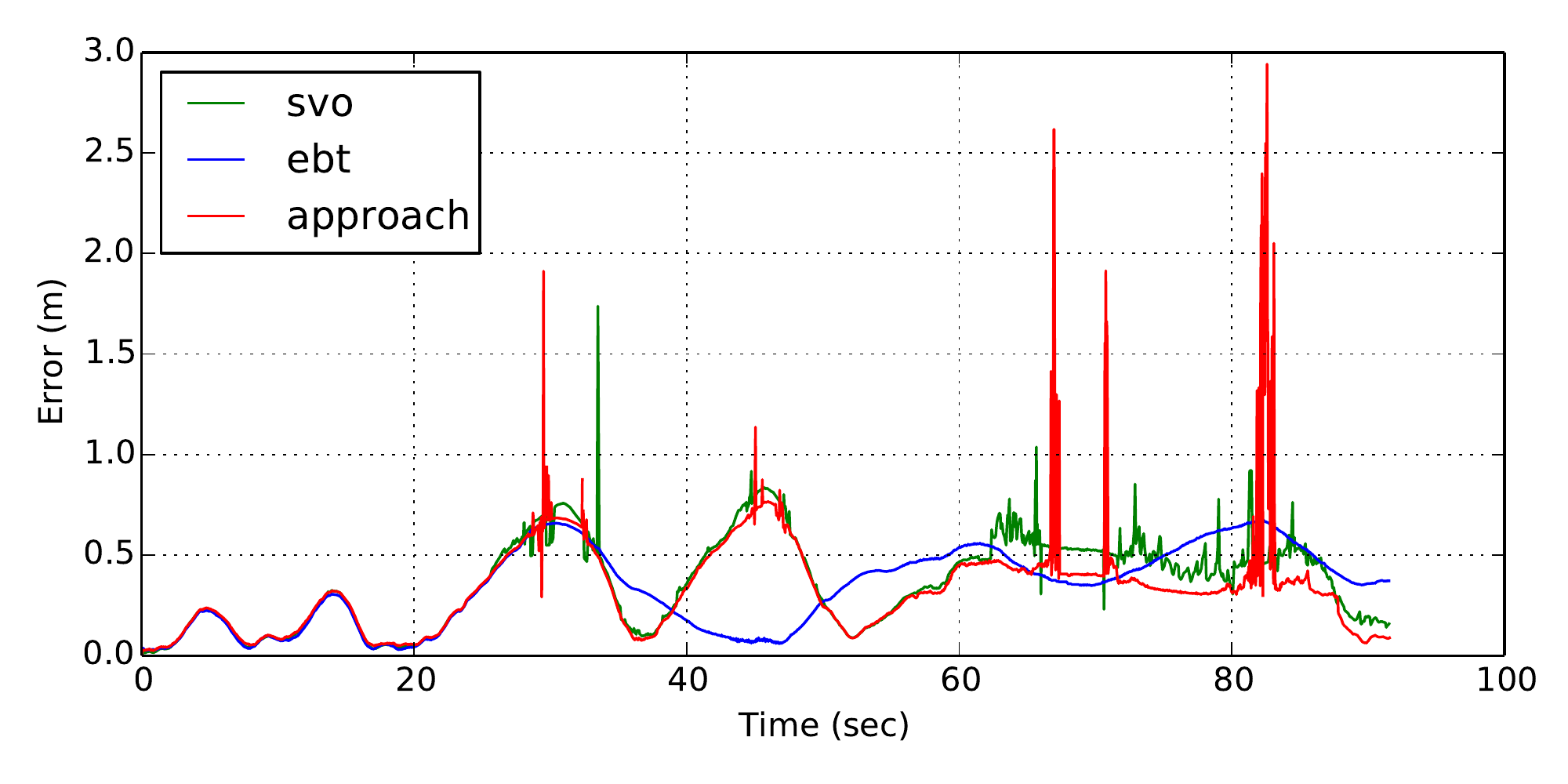}
        \end{subfigure}
        \begin{subfigure}[b]{0.33\linewidth}
                \includegraphics
                    [width=\linewidth]
                    {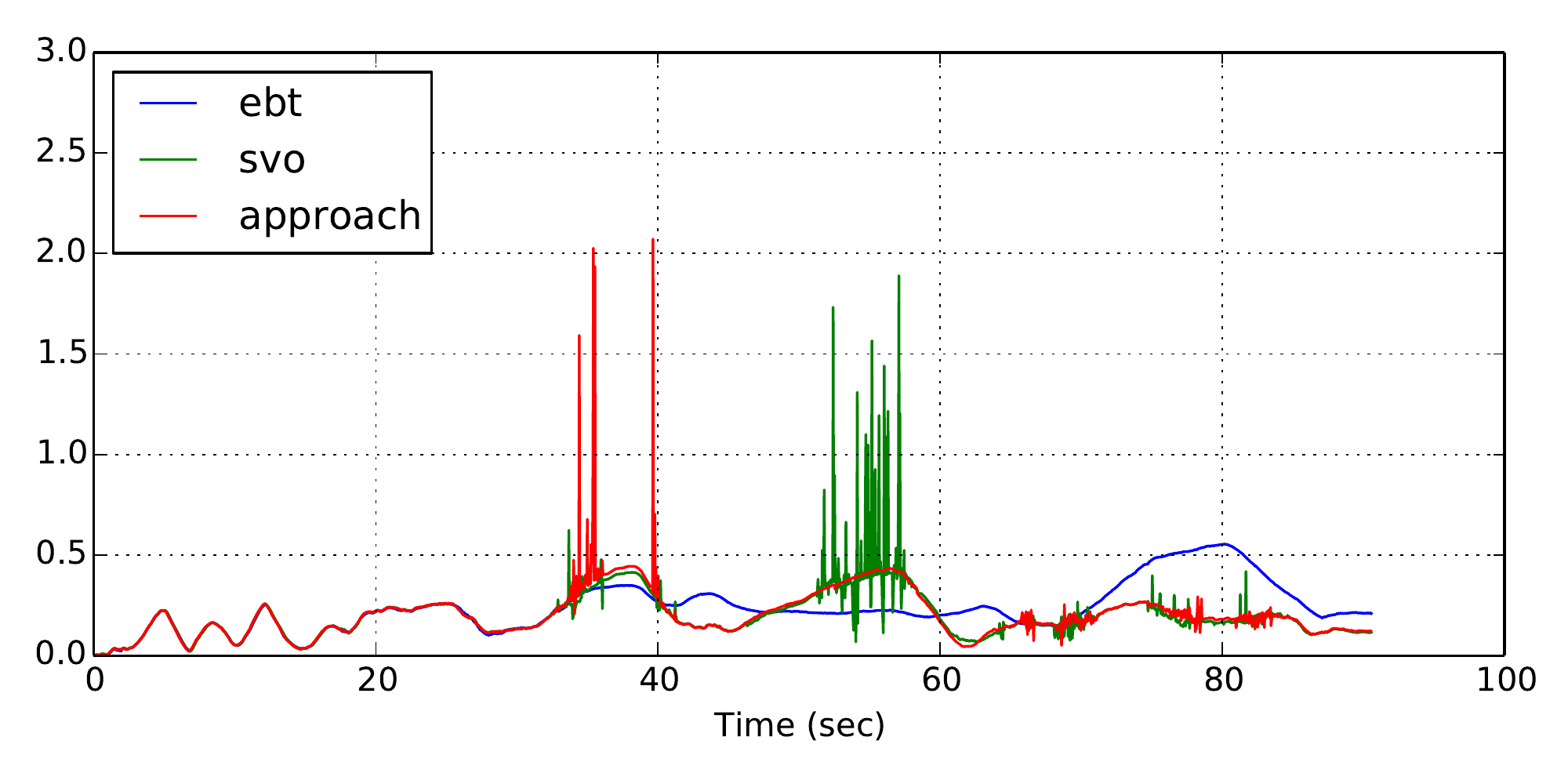}
        \end{subfigure}
        \begin{subfigure}[b]{0.33\linewidth}
                \includegraphics
                    [width=\linewidth]
                    {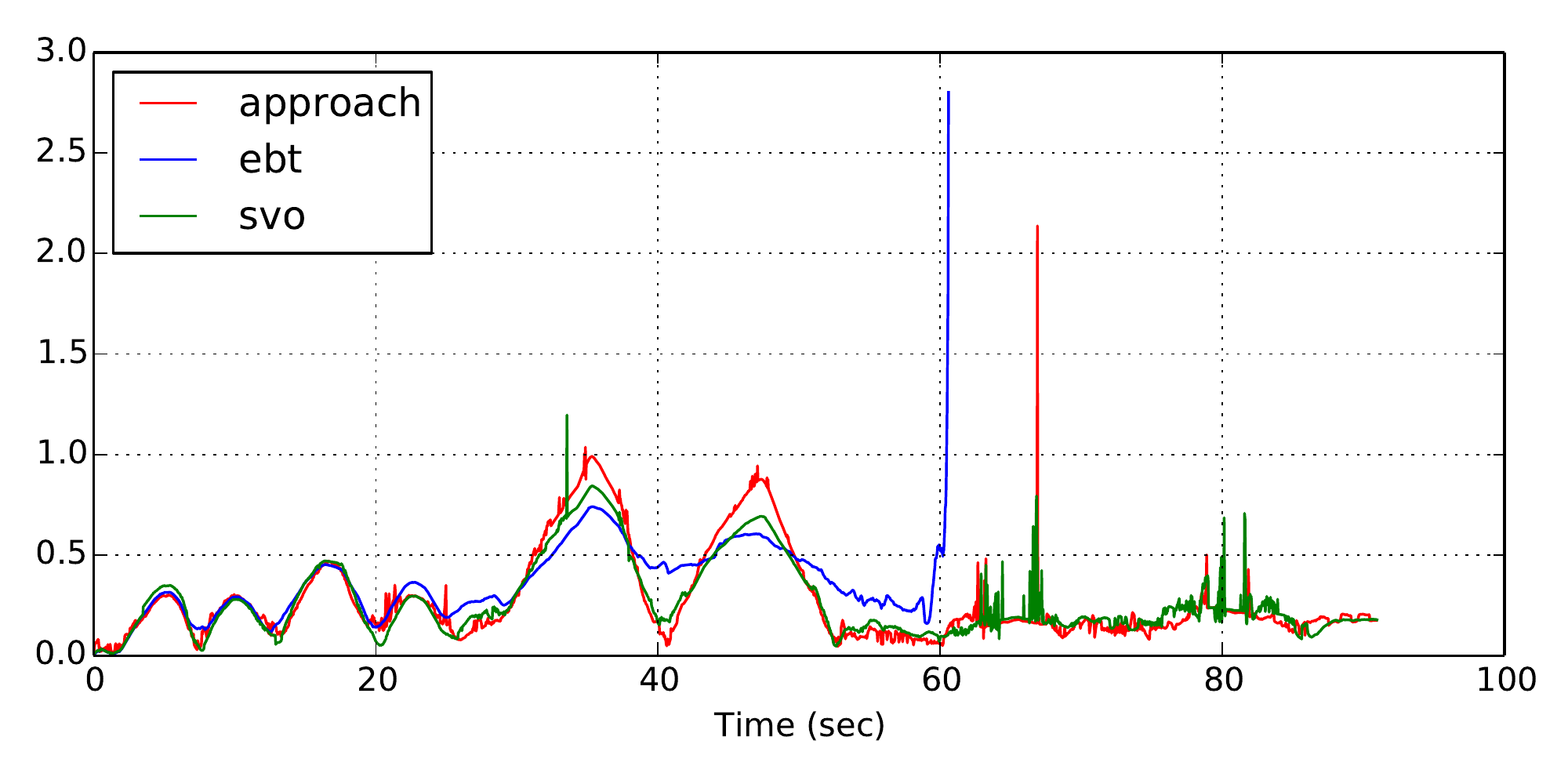}
        \end{subfigure}
    \caption{Per Frame Error of objects ( Left to Right: Orange Juice, Ronzoni Box and Tide Bottle) in 6 sequences. Figures show translational error in meters over time in seconds. First row shows tracking results as the camera is moved around a cluttered table, while the second row shows out of scene occlusion comparison caused by camera rotations and translations in front of tracked object. Note the initial spike in error for EBT particle tracking just before losing the object.}
    \label{fig:error_plots}
\end{figure*}


 We use ROS for implementation with the publicly available open source packages and using its communication framework for our system. This allows us to be modular and if required swap our current odometry and tracking systems for better methods. Available ROS package for SVO is used with modifications, our pose plugin subscribes to the relative pose measurement as a time stamped transformation from the odometry reference to the optical frame of the camera. The time dependent transform between the drifting odometry reference frame and the static world is one aspect that is continuously optimized and corrected for on-line. Additionally with our open source ROS package for 2D edge based tracking, we subscribe to the semantically labeled detections of objects. The health signal, along with the pose and covariance of the labeled detection is used by the object plugin in order to commit factors relating the optical frame and the objects reference frame using the open source OmniMapper ROS package. This time dependent sequence of detections is the second aspect continuously optimized. As shown in the flow diagram in Fig \ref{flowdiagram}, for the tracking plugin, this outer feedback results in resetting the tracked pose of the object using the graph estimate for rapid recovery, while barring further measurements of its label from contributing to the graph until reinitialisation is acknowledged. OmniMapper publishes ROS messages which are subscribed by the object based tracker for pose based reinitialization or if the counter for the feedback overflows. If the reported detections persist being erroneous from the current landmark pose for a given amount of frames or time, the object plugin permits the use of the tracker's own reinitialization method. Using OmniMapper we can subscribe to multiple objects with the same framework each being added individually to the graph with the messages being subscribed and published individually. 

\section{Experimental Results}
\label{sec:results}

We show our results on a challenging tabletop dataset for single and multiple objects. The current dataset consists of 6 challenging sequences with common household objects from the Amazon picking challenge and the PCL library. These standard objects allow easy access to their CAD models. The sequences have a high level of difficulty compared to standard smooth tracking datasets with partial and full occlusion of the object. Currently we show quantitative and qualitative results on 3 objects: Tide Box, Ronzoni Box and Orange Juice Carton. Each of the objects poses a different level of difficulty for tracking. Tide is largely textureless and smooth, making it tough to detect from various poses; Ronzoni box is textured and small making it difficult to track in large tabletop settings and Orange Juice Carton has both sharp and smooth edges with lot of texture leading to a high number of false data associations for detections and tracking. We plan to release the dataset with the code and add more objects to the library in the future.

The parameters we set for SVO are: Minimum Disparity to start tracking = 10 pixels, as we start close to the object, Maximum number of Keyframes = 100, to allow less overhead in batch optimization, Maximum number features tracked = 400, to allow fast addition of features in small spaces, Minimum number of features to be tracked = 20, for cases of textureless settings and having a short baseline of 5cm for adding new Keyframes. Other parameters are set to default settings of SVO ROS. All the parameters are kept constant for all the sequences in the dataset to allow a fair comparison. Parameters set for Object Tracking\cite{choi12:robus_euclid} are : Sampling step = 1cm, interval in which points are sampled from the CAD model, Number of particles = 20 and valid visible point threshold = 0.5, to allow high degree of partial occlusion. The parameters are fixed for all the sequences in the dataset. The noise parameter for the landmark factor is set to 15cm and 30 degrees and noise parameter for pose factor is set to 5 cm and 5 degrees along each axis for translation and rotation respectively.

We show quantitative results for per frame object tracking in Table 1. Here we measure the mean error per frame of a sequence with respect to ground truth pose of the object in the camera frame. We consider poses whose error is greater than 0.5m to be lost and do not consider them for our mean error computation. We show comparison to EBT\cite{choi12:robus_euclid} and EBT combined with SVO\cite{forster14:svo} but without the online update step. As can be seen from Table 1 we do better than EBT in all the 6 sequences for the per frame success $\%$  while EBT does better in mean error for the Ronzoni Box marginally  due to it being successful in only 20$\%$ of the sequence. Our approach does better than SVO both in terms of per frame success $\%$ and mean error while we found that odometry from SVO is very close to ground truth and does not suffer from drift. This leads us to believe that the online incremental feedback allows for more stable tracking. 

We plot the error over time for each sequence in Figure 4 with respect to the other 2 methods. As it can be seen from the figure, EBT fails for all the sequences midway while both SVO combined with EBT and our approach are able to track the object in the sequences. The spikes in the figures are the moments where EBT is about to fail but are reinitialized by feedback from Omnimapper. This shows that even in cases of full occlusion our method with feedback can relocalize and track the object successfully. This is shown in detail in the supplementary video.   

We also show quantitative results for error over the whole trajectory of the camera in each sequence. The comparison is done between SVO and our approach, with results being in Table 2. As it can be seen SVO does better than our approach at times, this is largely due to our use of ISAM2 which leads to large relinearization errors in case of wrong data associations. The mean error difference between the 2 approaches is less than 2 cm, showing that our method does not degrade to a large extent. Also another important finding is that in small spaces Object Based SLAM methods would not necessarily be beneficial over feature based methods in terms of localization error.  

\begin{table}[]
\centering
\begin{tabular}{ll|c|c|c|}
\cline{3-5}
                                                                                                 &        & \multicolumn{3}{c|}{\textbf{Method}}                                                                                                                                   \\ \hline
\multicolumn{2}{|c|}{\textbf{Sequence}}                                                                    & \textit{SVO}                                          & \textit{EBT}                                          & \textit{Approach}                                 \\ \hline
\multicolumn{1}{|l|}{\multirow{2}{*}{\begin{tabular}[c]{@{}l@{}}Orange\\ Juice\end{tabular}}}    & mean [m] & {\textbf{0.189}}  & 0.219  & 0.192  \\ \cline{2-2}
\multicolumn{1}{|l|}{}                                                                           & ratio \% & 30.10 & 21.95 & {\textbf{33.74}} \\ \hline
\multicolumn{1}{|l|}{\multirow{2}{*}{\begin{tabular}[c]{@{}l@{}}Orange \\ Occlud\end{tabular}}}  & mean [m] & {\textbf{0.233}}   & 0.257   & 0.250   \\ \cline{2-2}
\multicolumn{1}{|l|}{}                                                                           & ratio \% & 64.43  & 75.56  & {\textbf{82.99}}  \\ \hline
\multicolumn{1}{|l|}{\multirow{2}{*}{\begin{tabular}[c]{@{}l@{}}Ronzoni \\ Box\end{tabular}}}    & mean [m] & 0.303    & {\textbf{0.239}}    & 0.274    \\ \cline{2-2}
\multicolumn{1}{|l|}{}                                                                           & ratio \% & 41.37   & 23.23   & {\textbf{43.64}}   \\ \hline
\multicolumn{1}{|l|}{\multirow{2}{*}{\begin{tabular}[c]{@{}l@{}}Ronzoni \\ Occlud\end{tabular}}} & mean [m] & {\textbf{0.188}}     & 0.215     & 0.197     \\ \cline{2-2}
\multicolumn{1}{|l|}{}                                                                           & ratio \% & 99.09    & 94.08    & {\textbf{99.59}}    \\ \hline
\multicolumn{1}{|l|}{\multirow{2}{*}{\begin{tabular}[c]{@{}l@{}}Tide \\ Bottle\end{tabular}}}    & mean [m] & 0.289        & 0.288        & {\textbf{0.282}}        \\ \cline{2-2}
\multicolumn{1}{|l|}{}                                                                           & ratio \% & {\textbf{41.97}}       & 30.09       & 41.25       \\ \hline
\multicolumn{1}{|l|}{\multirow{2}{*}{\begin{tabular}[c]{@{}l@{}}Tide \\ Occlud\end{tabular}}}    & mean [m] & 0.210    & 0.293    & {\textbf{0.194}}    \\ \cline{2-2}
\multicolumn{1}{|l|}{}                                                                           & ratio \% & {\textbf{86.89}}   & 53.15   & 85.22   \\ \hline
\end{tabular}
\caption{Per Frame Translation Error less than 0.5 m. Ratio derived from frames with error under the threshold, while mean is then derived from corresponding frames.}
\label{my-label}
\end{table}

\begin{table}[]
\centering
\begin{tabular}{ll|c|c|}
\cline{3-4}
                                                                                                 &            & \multicolumn{2}{c|}{\textbf{Method}} \\ \hline
\multicolumn{2}{|c|}{\textbf{Sequence}}                                                                        & \textit{SVO}     & \textit{Approach} \\ \hline
\multicolumn{1}{|l|}{\multirow{2}{*}{\begin{tabular}[c]{@{}l@{}}Orange\\ Juice\end{tabular}}}    & mean [m]   & 0.090 $\pm$ 0.103  & 0.119 $\pm$ 0.094   \\ \cline{2-2}
\multicolumn{1}{|l|}{}                                                                           & median [m] & 0.063            & 0.094             \\ \hline
\multicolumn{1}{|l|}{\multirow{2}{*}{\begin{tabular}[c]{@{}l@{}}Orange \\ Occlud\end{tabular}}}  & mean [m]   & 0.095 $\pm$ 0.062  & 0.100 $\pm$ 0.060   \\ \cline{2-2}
\multicolumn{1}{|l|}{}                                                                           & median [m] & 0.073            & 0.083             \\ \hline
\multicolumn{1}{|l|}{\multirow{2}{*}{\begin{tabular}[c]{@{}l@{}}Ronzoni \\ Box\end{tabular}}}    & mean [m]   & 0.052 $\pm$ 0.028  & 0.117 $\pm$ 0.055   \\ \cline{2-2}
\multicolumn{1}{|l|}{}                                                                           & median [m] & 0.050            & 0.109             \\ \hline
\multicolumn{1}{|l|}{\multirow{2}{*}{\begin{tabular}[c]{@{}l@{}}Ronzoni \\ Occlud\end{tabular}}} & mean [m]   & 0.015 $\pm$ 0.008  & 0.038 $\pm$ 0.021   \\ \cline{2-2}
\multicolumn{1}{|l|}{}                                                                           & median [m] & 0.013            & 0.030             \\ \hline
\multicolumn{1}{|l|}{\multirow{2}{*}{\begin{tabular}[c]{@{}l@{}}Tide \\ Bottle\end{tabular}}}    & mean [m]   & 0.135 $\pm$ 0.045  & 0.115 $\pm$ 0.053   \\ \cline{2-2}
\multicolumn{1}{|l|}{}                                                                           & median [m] & 0.126            & 0.100             \\ \hline
\multicolumn{1}{|l|}{\multirow{2}{*}{\begin{tabular}[c]{@{}l@{}}Tide \\ Occlud\end{tabular}}}    & mean [m]   & 0.108 $\pm$ 0.103  & 0.124 $\pm$ 0.099   \\ \cline{2-2}
\multicolumn{1}{|l|}{}                                                                           & median [m] & 0.069            & 0.083             \\ \hline
\multicolumn{1}{|l|}{\multirow{2}{*}{\begin{tabular}[c]{@{}l@{}}All \end{tabular}}}            & mean [m]   & 0.083 $\pm$ 0.058  & 0.102 $\pm$ 0.064   \\ \cline{2-2}
\multicolumn{1}{|l|}{}                                                                           & median [m] & 0.066            & 0.083             \\ \hline
\end{tabular}
\caption{Absolute Trajectory error with respect to motion capture ground truth system. Shown is a comparison between our approach and SVO.}
\label{table:ground_truth_error}
\end{table}

\begin{figure*}
        \begin{subfigure}[b]{0.33\linewidth}
                \includegraphics
                    [width=\linewidth]
                    {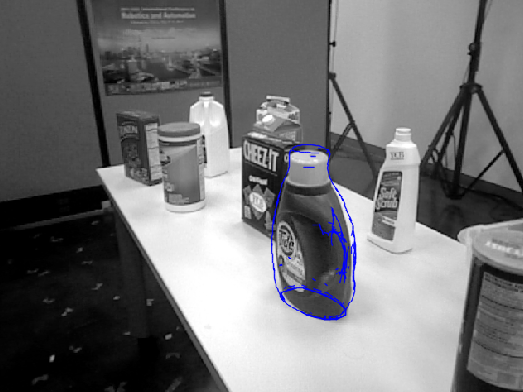}
        \end{subfigure}%
        \begin{subfigure}[b]{0.33\linewidth}
                \includegraphics
                    [width=\linewidth]
                    {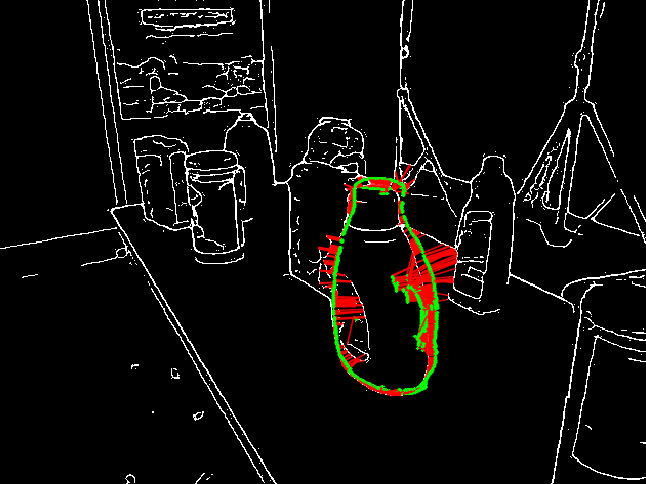}
        \end{subfigure}%
        \begin{subfigure}[b]{0.33\linewidth}
                \includegraphics
                    [width=\linewidth]
                    {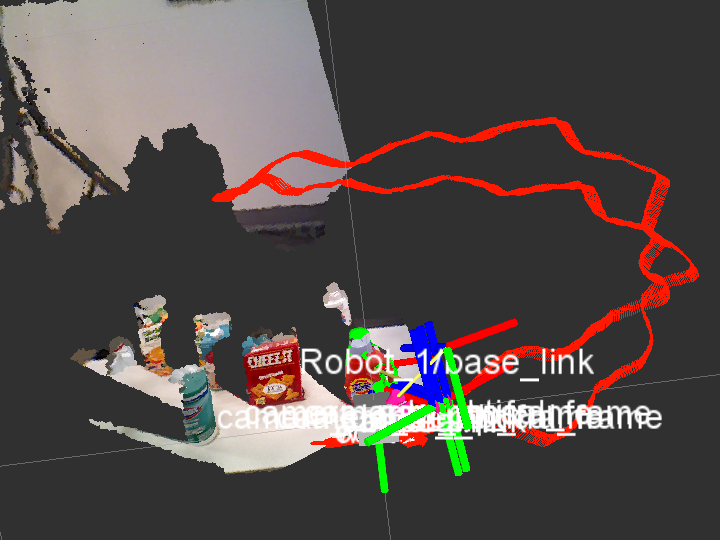}
        \end{subfigure}
        \begin{subfigure}[b]{0.33\linewidth}
                \includegraphics
                    [width=\linewidth]
                    {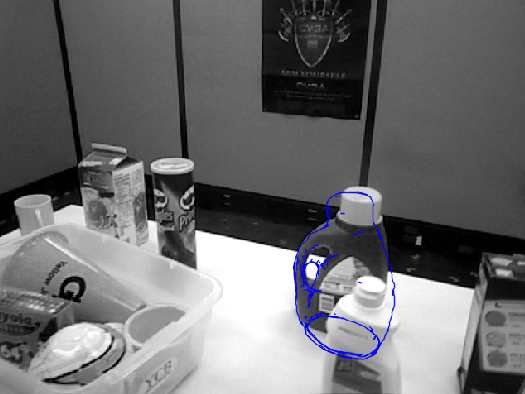}
        \end{subfigure}%
        \begin{subfigure}[b]{0.33\linewidth}
                \includegraphics
                    [width=\linewidth]
                    {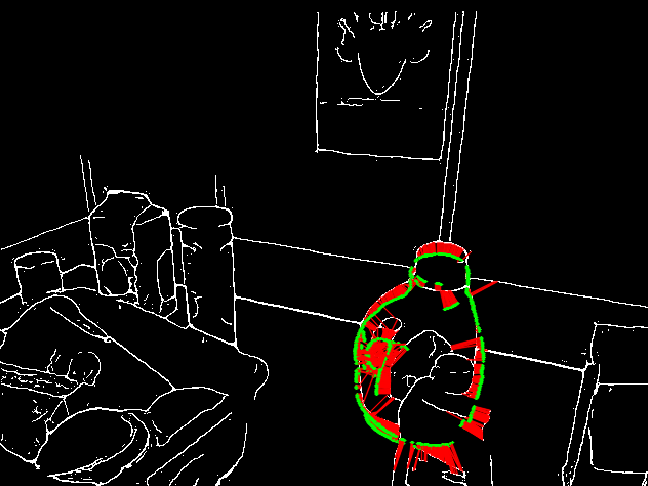}
        \end{subfigure}%
        \begin{subfigure}[b]{0.33\linewidth}
                \includegraphics
                    [width=\linewidth]
                    {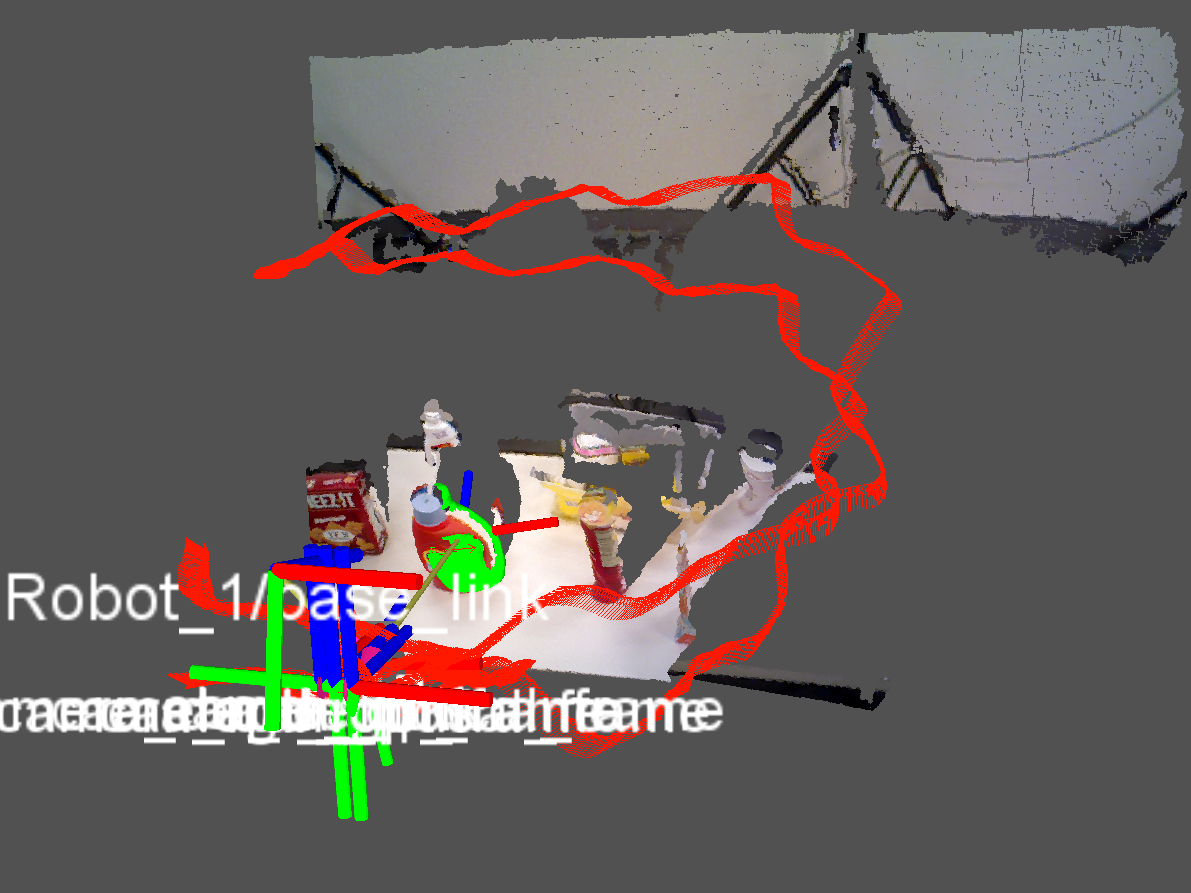}
        \end{subfigure}
    \caption{Above shown are frames from the Tide sequence (Left to Right): Tracked hypotheses superimposed on the image,  the edge based error using the select edges from the object model, and finally the optimized trajectory derived from the combined SVO and object measurements. Point cloud shown is only used as visual aid. Second row shows instances of foreground occlusion.}
    \label{fig:factorgraph}
\end{figure*}
We present some of our qualitative results in Fig 5. In Fig 5, we show results for the Tide sequence for the challenging case of partial occlusion and narrow viewing angle. As can be seen from the figure, the tracker is able to steadily track in both cases even in such difficult situations. The first and the second column show the current object pose projected in the current gray and edge image. The 3rd column shows the current trajectory of the camera with object overlaid on the current scene. It can be seen that even in large room cases our method is able to track efficiently.

\begin{figure}
        \begin{subfigure}[b]{0.5\linewidth}
                \includegraphics
                    [width=\linewidth]
                    {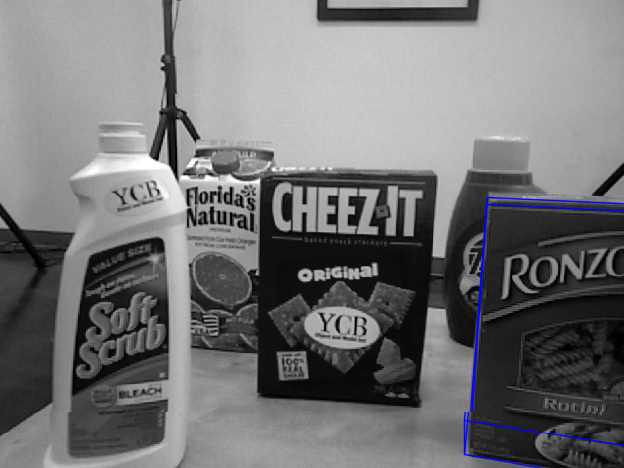}
        \end{subfigure}%
        \begin{subfigure}[b]{0.5\linewidth}
                \includegraphics
                    [width=\linewidth]
                    {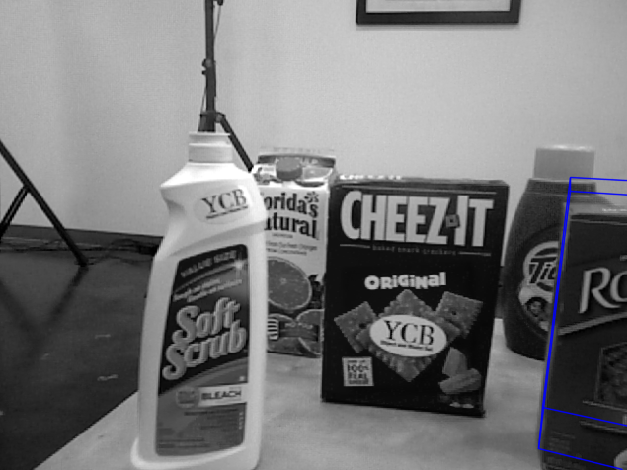}
        \end{subfigure}
        \begin{subfigure}[b]{0.5\linewidth}
                \includegraphics
                    [width=\linewidth]
                    {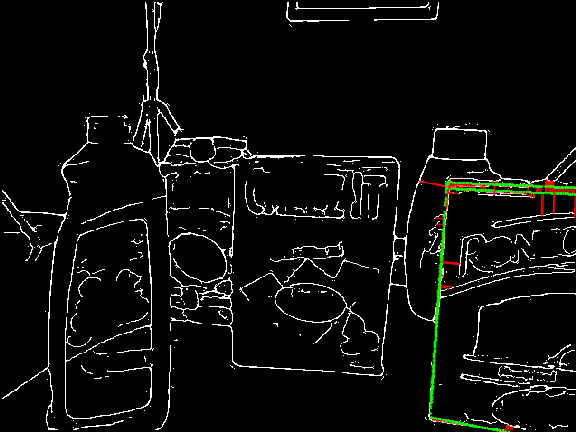}
        \end{subfigure}%
        \begin{subfigure}[b]{0.5\linewidth}
                \includegraphics
                    [width=\linewidth]
                    {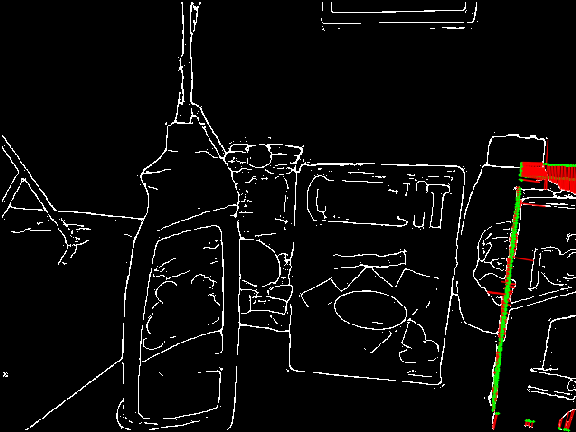}
        \end{subfigure}
    \caption{Top: Object superimposed on the image , Bottom: Edge image superimposed with object. Shown above on the left is the view of the object in the Ronzoni Occlud sequence just before exiting the scene. On the right is the relocalization of the object as it comes back into view.}
    \label{fig:ron_occ}
\end{figure}

\begin{figure}
        \begin{subfigure}[b]{0.5\linewidth}
                \includegraphics
                    [width=\linewidth]
                    {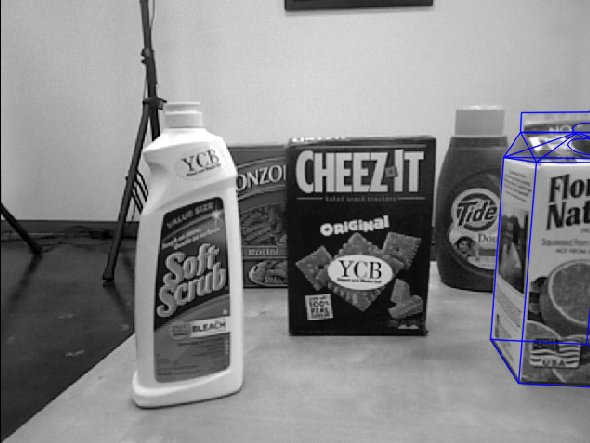}
        \end{subfigure}%
        \begin{subfigure}[b]{0.5\linewidth}
                \includegraphics
                    [width=\linewidth]
                    {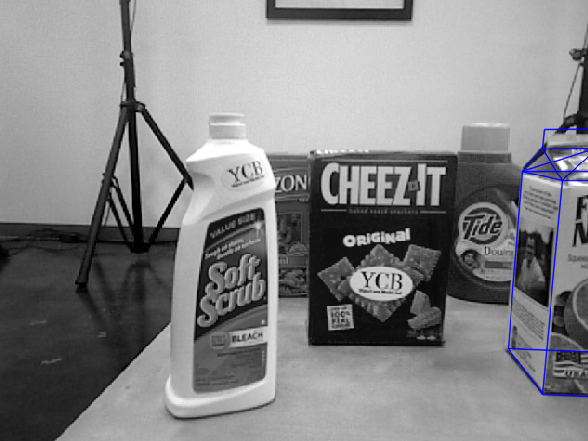}
        \end{subfigure}
        \begin{subfigure}[b]{0.5\linewidth}
                \includegraphics
                    [width=\linewidth]
                    {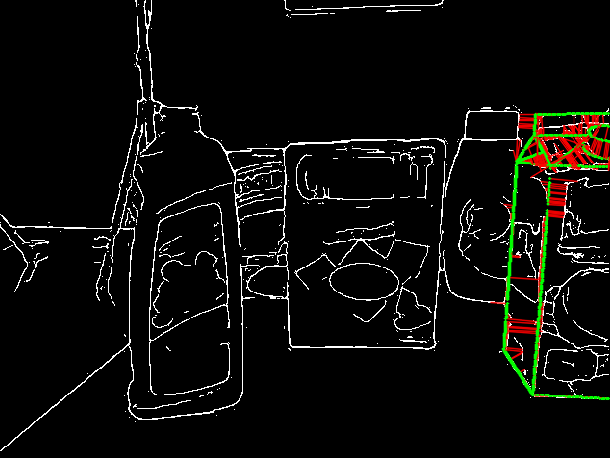}
        \end{subfigure}%
        \begin{subfigure}[b]{0.5\linewidth}
                \includegraphics
                    [width=\linewidth]
                    {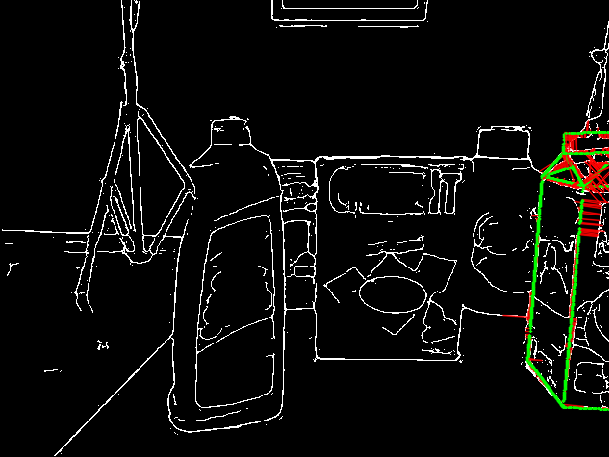}
        \end{subfigure}
    \caption{Top: Object superimposed on the image , Bottom: Edge image superimposed with object. Similar to Fig \ref{fig:ron_occ}, left side shows the Orange Occlud sequence just before the object exits the scene, while the right hand side shows the object being still tracked with high occlusion.}
    \label{fig:oj_occ}
\end{figure}
In Fig 6 \& Fig 7, we show a sequence of 2 images from Ronzoni Occluded Sequence and Orange Juice Occluded Sequence. We show how in consecutive frames even after more than 50\% occlusion of these objects our tracker is able to track them effectively and relocalize an object by only seeing a part of it. This shows our tracker is highly robust to occlusion and the semantic map allows us to relocalize objects without any adhoc detection. More qualitative results are presented in the supplementary video.



\section{Conclusions and Future Work}
\label{sec:conclusions}

In this work we have demonstrated methods of improving object tracking reinitialization for monocular cameras by means of incorporating traditional tracking algorithms into a higher-level framework permitting modular measurement fusion, including position feedback from both visual odometry and semantic mapping uncertainties. In this manner, alternative tracking methods and motion estimation can be combined to generate sparse graph representations of world environments, reducing optimization problems for small memory and computationally limited applications, such as small mobile robotic platforms.

Future work within this domain could include evaluating the performance of various combinations of other available VO and object trackers, or perhaps collective of a multiple of each type, and charting the effects of different object types under degrees of occlusion other than for monocular imagery. Additionally, associating an object with its own pose-chain when recognized as non-static could provide a relatively simple extension of the current information sharing framework for dynamic environments, as VO with accompanying objects could be used to discern the relatively dynamic elements in a scene. However, perhaps the most promising extension of this framework would be for object learning, thus moving away from priori model dependent tracking.



\bibliographystyle{IEEEtran}
\bibliography{main}

\begin{thebibliography}{10}
\providecommand{\url}[1]{#1}
\csname url@samestyle\endcsname
\providecommand{\newblock}{\relax}
\providecommand{\bibinfo}[2]{#2}
\providecommand{\BIBentrySTDinterwordspacing}{\spaceskip=0pt\relax}
\providecommand{\BIBentryALTinterwordstretchfactor}{4}
\providecommand{\BIBentryALTinterwordspacing}{\spaceskip=\fontdimen2\font plus
\BIBentryALTinterwordstretchfactor\fontdimen3\font minus
  \fontdimen4\font\relax}
\providecommand{\BIBforeignlanguage}[2]{{%
\expandafter\ifx\csname l@#1\endcsname\relax
\typeout{** WARNING: IEEEtran.bst: No hyphenation pattern has been}%
\typeout{** loaded for the language `#1'. Using the pattern for}%
\typeout{** the default language instead.}%
\else
\language=\csname l@#1\endcsname
\fi
#2}}
\providecommand{\BIBdecl}{\relax}
\BIBdecl

\bibitem{trevor14:omnimapper}
A.~J. Trevor, J.~G. Rogers~III, and H.~I. Christensen, ``Omnimapper: A modular
  multimodal mapping framework,'' in \emph{Proceedings on the IEEE
  International Conference on Robotics and Automation (ICRA)}.\hskip 1em plus
  0.5em minus 0.4em\relax IEEE, 2014, pp. 1983--1990.

\bibitem{klein2007:PTAM}
G.~Klein and D.~Murray, ``Parallel tracking and mapping for small ar
  workspaces,'' in \emph{Mixed and Augmented Reality, 2007. ISMAR 2007. 6th
  IEEE and ACM International Symposium on}, Nov 2007, pp. 225--234.

\bibitem{PTAM2}
------, ``Parallel tracking and mapping for small ar workspaces,'' in
  \emph{Mixed and Augmented Reality, 2007. ISMAR 2007. 6th IEEE and ACM
  International Symposium on}, Nov 2007, pp. 225--234.

\bibitem{newcombe2011kinectfusion}
R.~A. Newcombe, S.~Izadi, O.~Hilliges, D.~Molyneaux, D.~Kim, A.~J. Davison,
  P.~Kohi, J.~Shotton, S.~Hodges, and A.~Fitzgibbon, ``Kinectfusion: Real-time
  dense surface mapping and tracking,'' in \emph{Mixed and augmented reality
  (ISMAR), 2011 10th IEEE international symposium on}.\hskip 1em plus 0.5em
  minus 0.4em\relax IEEE, 2011, pp. 127--136.

\bibitem{MorenoNSKD13}
R.~F. Salas{-}Moreno, R.~A. Newcombe, H.~Strasdat, P.~H.~J. Kelly, and A.~J.
  Davison, ``{SLAM++:} simultaneous localisation and mapping at the level of
  objects,'' in \emph{2013 {IEEE} Conference on Computer Vision and Pattern
  Recognition, Portland, OR, USA, June 23-28, 2013}, 2013, pp. 1352--1359.

\bibitem{bao2011semantic}
S.~Y. Bao and S.~Savarese, ``Semantic structure from motion,'' in
  \emph{Computer Vision and Pattern Recognition (CVPR), 2011 IEEE Conference
  on}.\hskip 1em plus 0.5em minus 0.4em\relax IEEE, 2011, pp. 2025--2032.

\bibitem{krull2014:6dof}
``6-dof model based tracking via object coordinate regression,'' in
  \emph{Computer Vision -- ACCV 2014}, ser. Lecture Notes in Computer Science,
  D.~Cremers, I.~Reid, H.~Saito, and M.-H. Yang, Eds., 2015, vol. 9006.

\bibitem{lu2014:unsupervised}
``Unsupervised dense object discovery, detection, tracking and
  reconstruction,'' in \emph{Computer Vision – ECCV 2014}, ser. Lecture Notes
  in Computer Science, D.~Fleet, T.~Pajdla, B.~Schiele, and T.~Tuytelaars,
  Eds., 2014, vol. 8690.

\bibitem{forster14:svo}
C.~Forster, M.~Pizzoli, and D.~Scaramuzza, ``Svo: Fast semi-direct monocular
  visual odometry,'' in \emph{Robotics and Automation (ICRA), 2014 IEEE
  International Conference on}, May 2014, pp. 15--22.

\bibitem{LSDslam}
\BIBentryALTinterwordspacing
J.~Engel, T.~Sch{\"{o}}ps, and D.~Cremers, ``{LSD-SLAM:} large-scale direct
  monocular {SLAM},'' in \emph{Computer Vision - {ECCV} 2014 - 13th European
  Conference, Zurich, Switzerland, September 6-12, 2014, Proceedings, Part
  {II}}, 2014, pp. 834--849. [Online]. Available:
  \url{http://dx.doi.org/10.1007/978-3-319-10605-2-54}
\BIBentrySTDinterwordspacing

\bibitem{epnp}
\BIBentryALTinterwordspacing
V.~Lepetit, F.~Moreno{-}Noguer, and P.~Fua, ``Ep\emph{n}p: An accurate
  \emph{O}(\emph{n}) solution to the p\emph{n}p problem,'' \emph{International
  Journal of Computer Vision}, vol.~81, no.~2, pp. 155--166, 2009. [Online].
  Available: \url{http://dx.doi.org/10.1007/s11263-008-0152-6}
\BIBentrySTDinterwordspacing

\bibitem{choi12:robus_euclid}
C.~Choi and H.~Christensen, ``Robust 3d visual tracking using particle
  filtering on the special euclidean group: A combined approach of keypoint and
  edge features,'' \emph{The International Journal of Robotics Research},
  vol.~31, no.~4, pp. 498--519, 2012.

\bibitem{choi10:realtime}
------, ``Real-time 3d model-based tracking using edge and keypoint features
  for robotic manipulation,'' in \emph{Robotics and Automation (ICRA), 2010
  IEEE International Conference on}, May 2010, pp. 4048--4055.

\bibitem{gtsam}
F.~Dellaert, ``Factor {G}raphs and {GTSAM}: {A} {H}ands-on {I}ntroduction,''
  Georgia Tech, Tech. Rep. GT-RIM-CP\&R-2012-002, September 2012.

\bibitem{Kaess12ijrr}
M.~Kaess, H.~Johannsson, R.~Roberts, V.~Ila, J.~J. Leonard, and F.~Dellaert,
  ``isam2: Incremental smoothing and mapping using the bayes tree,'' \emph{The
  International Journal of Robotics Research}, vol.~31, no.~2, pp. 216--235,
  2012.

\bibitem{jcbb}
\BIBentryALTinterwordspacing
J.~Neira and J.~D. Tard{\'{o}}s, ``Data association in stochastic mapping using
  the joint compatibility test,'' \emph{{IEEE} T. Robotics and Automation},
  vol.~17, no.~6, pp. 890--897, 2001. [Online]. Available:
  \url{http://dx.doi.org/10.1109/70.976019}
\BIBentrySTDinterwordspacing

\end{thebibliography}

\end{document}